\documentclass[runningheads]{llncs}
\usepackage[T1]{fontenc}
\usepackage{graphicx}
\usepackage{booktabs}
\usepackage{makecell}
\usepackage{lipsum}
\usepackage{amssymb}
\usepackage{amsmath}
\usepackage[colorlinks,citecolor=blue,linkcolor=blue]{hyperref}
\usepackage{marvosym}
\usepackage{floatrow}
\newfloatcommand{capbtabbox}{table}[][]
\begin{document}
\title{Additional Positive Enables Better Representation Learning for Medical Images}
\titlerunning{Additional Positive Enables Better Representation Learning}
%
\author{Dewen Zeng\inst{1}$^{(\textrm{\Letter})}$ \and
Yawen Wu\inst{2} \and
Xinrong Hu\inst{1} \and
Xiaowei Xu\inst{3} \and 
Jingtong Hu\inst{2} \and 
Yiyu Shi\inst{1}$^{(\textrm{\Letter})}$
}
\authorrunning{D. Zeng, et al.}
%

%
\institute{
University of Notre Dame, Notre Dame, IN, USA \\\email{\{dzeng2, yshi4\}@nd.edu}\and
University of Pittsburgh, Pittsburgh, PA, USA \and
Guangdong Provincial People's Hospital, Guangzhou, China
}
%
%
%
\maketitle              
\begin{abstract}
This paper presents a new way to identify additional positive pairs for BYOL, a state-of-the-art (SOTA) self-supervised learning framework, to improve its representation learning ability. Unlike conventional BYOL which relies on only one positive pair generated by two augmented views of the same image, we argue that information from different images with the same label can bring more diversity and variations to the target features, thus benefiting representation learning. To identify such pairs without any label, we investigate TracIn, an instance-based and computationally efficient influence function, for BYOL training. Specifically, TracIn is a gradient-based method that reveals the impact of a training sample on a test sample in supervised learning. We extend it to the self-supervised learning setting and propose an efficient batch-wise per-sample gradient computation method to estimate the pairwise TracIn for representing the similarity of samples in the mini-batch during training. For each image, we select the most similar sample from other images as the additional positive and pull their features together with BYOL loss. Experimental results on two public medical datasets (i.e., ISIC 2019 and ChestX-ray) demonstrate that the proposed method can improve the classification performance compared to other competitive baselines in both semi-supervised and transfer learning settings. 


\end{abstract}
\section{Introduction}

Self-supervised learning (SSL) has been extremely successful in learning good image representations without human annotations for medical image applications like classification \cite{sowrirajan2021moco,azizi2021big,zhang2022contrastive} and segmentation \cite{bai2019self,chaitanya2020contrastive,hu2021semi}. Usually, an encoder is pre-trained on a large-scale unlabeled dataset. Then, the pre-trained encoder is used for efficient training on downstream tasks with limited annotation \cite{tajbakhsh2020embracing,jaiswal2020survey}.
Recently, contrastive learning has become the state-of-the-art (SOTA) SSL method due to its powerful learning ability. A naive contrastive learning method learns by pulling the representations of different augmented views of the same image (a.k.a positive pair) together and pushing the representation of different images (a.k.a negative pair) apart \cite{chen2020simple}. 
The main disadvantage of this method is its heavy reliance on negative pairs, making it necessary to use a large batch size \cite{chen2020simple} or memory banks \cite{he2020momentum} to ensure effective training.
To overcome this challenge, BYOL \cite{grill2020bootstrap} proposes two siamese neural networks - the online and target networks. 
The online network is trained to predict the target network representation of the same image under a different augmented view, requiring only one positive pair per sample. This approach makes BYOL more resilient to batch size and the choice of data augmentations.

As the positive pair in BYOL is generated from the same image, the diversity of features within the positive pair could be quite limited. For example, one skin disease may manifest differently in different patients or locations, but such information is often overlooked in the current BYOL framework.
In this paper, we argue that such feature diversity can be increased by adding additional positive pairs from other samples with the same label (a.k.a. True Positives). Identifying such pairs without human annotation is challenging because of the unrelated information in medical images, such as the background pathology. One straightforward way is to use feature similarity: two images are considered positive if their representations are close to each other in the feature space.
However, samples with different labels might also be close in the feature space because the learned encoder is not perfect. Considering them as positive might further pull them together after learning, leading to degraded performance.

To solve this problem, we propose BYOL-TracIn, which improves vanilla BYOL using the TracIn influence function. 
Instead of quantifying the similarity of two samples based on feature similarity, we propose using TracIn to estimate their similarity by calculating the impact of training one sample on the other. 
TracIn \cite{pruthi2020estimating} is a gradient-based influence function that measures the loss reduction of one sample by the training process of another sample. Directly applying TracIn in BYOL is non-trivial as it requires the gradient of each sample and careful selection of model checkpoints and data augmentations to accurately estimate sample impacts without labels. To avoid per-sample gradient computation, we introduce an efficient method that computes the pairwise TracIn in a mini-batch with only one forward pass. For each image in the mini-batch, the sample from other images with the highest TracIn values is selected as the additional positive pair. Their representation distance is then minimized using BYOL loss. To enhance positive selection accuracy, we propose to use a pre-trained model for pairwise TracIn computation as it focuses more on task-related features compared to an on-the-fly model. Light augmentations are used on the samples for TracIn computation to ensure stable positive identification.
To the best of our knowledge, we are the first to incorporate additional positive pairs from different images in BYOL. 
Our extensive empirical results show that our proposed method outperforms other competing approaches in both semi-supervised and transfer learning settings for medical image classification tasks.











\section{Related Work}
\noindent\textbf{Self-supervised Learning.} 
Most SSL methods can be categorized as either generative \cite{doersch2015unsupervised,zhang2016colorful} or discriminative \cite{noroozi2016unsupervised,gidaris2018unsupervised}, in which pseudo labels are automatically generated from the inputs. Recently, contrastive learning \cite{chen2020simple,he2020momentum,zbontar2021barlow} as a new discriminative SSL method has dominated this field because of its excellent performance. SimCLR \cite{chen2020simple} and MoCo \cite{he2020momentum} are two typical contrastive learning methods that try to attract positive pairs and repulse negative pairs. However, these methods rely on a large number of negative samples to work well. BYOL \cite{grill2020bootstrap} improves contrastive learning by directly predicting the representation output from another view and achieves SOTA performance. As such, only positive pairs are needed for training. 
SimSiam \cite{chen2021exploring} further proves that stop-gradient plays an essential role in the learning stability of siamese neural networks.
Since the positive pairs in BYOL come from the same image, the feature diversity from different images of the same label is ignored. Our method introduces a novel way to accurately identify such positive pairs and attract them in the feature space.

\noindent\textbf{Influence Function.} 
The influence function (IF) was first introduced to machine learning models in \cite{koh2017understanding} to study the following question: which training points are most responsible for a given prediction? Intuitively, if we remove an important sample from the training set, we will get a large increase in the test loss. IF can be considered as an interpretability score that measures the importance of all training samples on the test sample. Aside from IF, other types of scores and variants have also been proposed in this field \cite{hara2019data,chen2020multi,barshan2020relatif}. Since IF is extremely computationally expensive, TracIn \cite{pruthi2020estimating} was proposed as an efficient alternative to estimate training sample influence using first-order approximation. Our method extends the normal TracIn to the SSL setting (i.e., BYOL) with a sophisticated positive pair selection schema and an efficient batch-wise per-sample gradient computation method, demonstrating that aside from model interpretation, TracIn can also be used to guide SSL pre-training.

\section{Method}

\subsection{Framework Overview}

\begin{figure}[t]
\centering
  \includegraphics[width=1.0\textwidth]{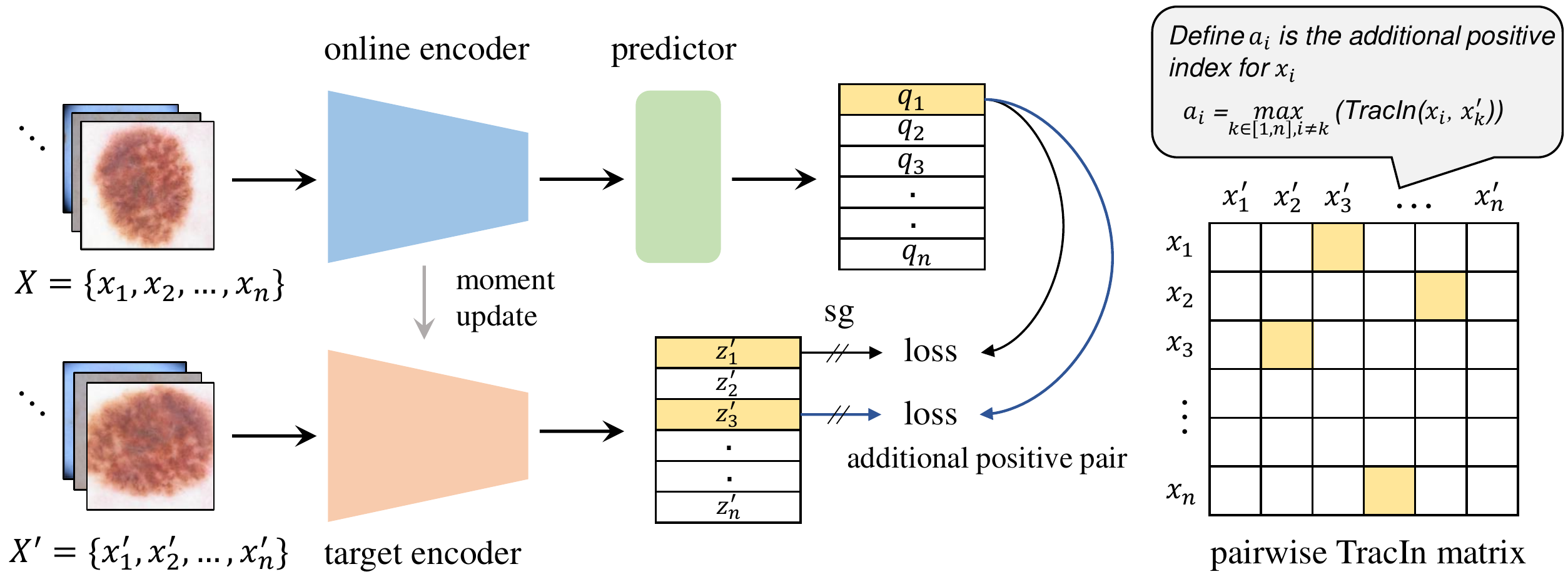}
\caption{Overview of the proposed BYOL-TracIn framework. $X$ and $X^\prime$ represent two augmentations of the mini-batch inputs. BYOL-TracIn minimizes the similarity loss of two views of the same image (e.g., $q_1$ and $z_1^\prime$) as well as the similarity loss of the additional positive (e.g., $z_3^\prime$) identified by our TracIn algorithm. sg means stop-gradient.}
\label{fig:overview}
\end{figure}

Our BYOL-TracIn framework is built upon classical BYOL method \cite{grill2020bootstrap}. Fig. \ref{fig:overview} shows the overview of our framework. Here, we use $x_1$ as the anchor sample for an explanation, and the same logic can be applied to all samples in the mini-batch. Unlike classical BYOL where only one positive pair ($x_1$ and $x_1^\prime$) generated from the same image is utilized, we use the influence function, TracIn, to find another sample ($x_3^\prime$) from the batch that has the largest impact on the anchor sample. During training, the representations distance of $x_1$ and $x_3^\prime$ will also be minimized. We think this additional positive pair can increase the variance and diversity of the features of the same label, leading to better clustering in the feature space and improved learning performance. The pairwise TracIn matrix is computed using first-order gradient approximation which will be discussed in the next section. For simplicity, this paper only selects the top-1 additional sample, 
but our method can be easily extended to include top-k ($k>1$) additional samples.

\subsection{Additional Positive Selection using TracIn}


\noindent\textbf{Idealized TracIn and Its First-order Approximation.}
Suppose we have a training dataset $D=\{x_1, x_2, ..., x_n\}$ with $n$ samples. $f_w(\cdot)$ is a model with parameter $w\in \mathbb{R}$, and $\ell(w,x_i)$ is the loss function when model parameter is $w$ and training example is $x_i$. The training process in iteration $t$ can be viewed as minimizing the training loss $\ell(w_t,x_t)$ and updating parameter $w_t$ to $w_{t+1}$ using gradient descent (suppose only $x_t \in D$ is used for training in each iteration). Then the idealized TracIn of one sample $x_i$ on another sample $x_k$ can be defined as the total loss reduction by training $x_i$ in the whole training process.
\begin{equation}
    \text{TracInIdeal}(x_i,x_k) = \sum_{t: x_t=x_i}^T(\ell(w_t,x_k) - \ell(w_{t+1},x_k)).
\end{equation}
where $T$ is the total number of iterations. If stochastic gradient descent is utilized as the optimization method, we can approximately express the loss reduction after iteration $t$ as $\ell(w_{t+1}, x_k)-\ell(w_{t}, x_k)=\triangledown\ell(w_{t}, x_k)\cdot(w_{t+1}-w_t) + O(||\Delta w_t||^2)$. The parameter change in iteration $t$ is $\Delta w_t=w_{t+1}-w_t=-\eta_t \triangledown\ell(w_{t}, x_t)$, in which $\eta_t$ is the learning rate in iteration $t$, and $x_t$ is the training example. Since $\eta_t$ is usually small during training, we can ignore the high order term $O(||\Delta w_t||^2)$, and the first-order TracIn can be formulated as:
\begin{equation}
    \text{TracIn}(x_i,x_k) = \sum_{t: x_t=x_i}^T\eta_t \triangledown\ell(w_{t}, x_k)\cdot  \triangledown\ell(w_{t}, x_i).
\label{equa:tracin}
\end{equation}

The above equation reveals that we can estimate the influence of $x_i$ on $x_k$ by summing up their gradient dot products across all training iterations. In practical BYOL training, the optimization is usually done on mini-batches, and it is impossible to save the gradients of a sample for all iterations. However, we can use the TracIn of \textbf{the current iteration} to represent the similarity of two samples in the mini-batch because we care about the pairwise relative influences instead of the exact total values across training. Intuitively, if the TracIn of two samples is large in the current iteration, this means that the training of one sample can benefit the other sample a lot because they share some common features. Therefore, they are similar to each other.

\noindent\textbf{Efficient Batch-wise TracIn Computation.}
Equation \ref{equa:tracin} requires the gradient of each sample in the mini-batch for pairwise TracIn computation. However, it is prohibitively expensive to compute the gradient of samples one by one. Moreover, calculating the dot product of gradients on the entire model is computationally and memory-intensive, especially for large deep-learning models where there could be millions or trillions of parameters. Therefore, we work with the gradients of the last linear layer in the online predictor.

As current deep learning frameworks (e.g., Pytorch and TensorFlow) do not support per-sample gradient when the batch size is larger than 1, we use the following method to efficiently compute the per-sample gradient of the last layer. Suppose the weight matrix of the last linear layer is $W\in \mathbb{R}^{m\times n}$, where $m$ and $n$ are the numbers of input and output units. $f(q)=2-2\cdot \langle q, z \rangle/ (\lVert q \rVert_2 \cdot \lVert z \rVert_2)$ is the standard BYOL loss function, where $q$ is the online predictor output (a.k.a., logits) and $z$ is the target encoder output that can be viewed as a constant during training. We have $q=Wa$, where $a$ is the input to the last linear layer. According to the chain rule, the gradient of the last linear layer can be computed as $\triangledown_Wf(q) = \triangledown_qf(q) a^T$, in which the gradient of the logits can be computed by:
\begin{equation}
    \triangledown_qf(q) = 2\cdot (\frac{\langle q, z \rangle \cdot q}{\lVert q \rVert_2^3 \cdot \lVert z \rVert_2} - \frac{z}{\lVert q \rVert_2 \cdot \lVert z \rVert_2}).
\label{equa:delta_logits}
\end{equation}
Therefore, the TracIn of sample $x_i$ and $x_k$ at iteration $t$ can be computed as:
\begin{equation}
\begin{split}
    \text{TracIn}(x_i,x_k) & \approx \eta_t \triangledown_Wf(q_i) \cdot \triangledown_Wf(q_k) \\
    & = \eta_t (\triangledown_qf(q_i) \cdot \triangledown_qf(q_k))(a_i\cdot a_k).
\end{split}
\label{equa:effcient_tracin}
\end{equation}

Equation \ref{equa:delta_logits} and \ref{equa:effcient_tracin} tell us that the per-sample gradient of the last linear layer can be computed by using the inputs of this layer and the gradient of the output logits for each sample, which can be achieved with only one forward pass on the mini-batch. This technique greatly speeds up the TracIn computation and makes it possible to be used in BYOL.

\noindent\textbf{Using Pre-trained Model to Increase True Positives.}
During the pre-training stage of BYOL, especially in the early stages, the model can be unstable and may focus on unrelated features in the background instead of the target features. This can result in the selection of wrong positive pairs while using TracIn.
For example, the model may identify all images with skin diseases on the face as positive pairs, even if they are from different diagnostics, as it focuses on the face feature instead of the diseases.
To address this issue, we suggest using a pre-trained model to select additional positives with TracIn to guide BYOL training. This is because a pre-trained model is more stable and well-trained to focus on the target features, thus increasing the selected true positive ratio.

\section{Experiments and Results}

\subsection{Experimental Setups}

\noindent\textbf{Datasets.}
We evaluate the performance of the proposed BYOL-TracIn on four publicly available medical image datasets.
\textbf{(1) ISIC 2019 dataset} is a dermatology dataset that contains 25,331 dermoscopic images among nine different diagnostic categories \cite{codella2019skin,tschandl2018ham10000,combalia2019bcn20000}.
\textbf{(2) ISIC 2016 dataset} was hosted in ISBI 2016 \cite{gutman2016skin}. It contains 900 dermoscopic lesion images with two classes benign and malignant.
\textbf{(3) ChestX-ray dataset} is a chest X-ray database that comprises 108,948 frontal view X-ray images of 32,717 unique patients with 14 disease labels \cite{wang2017chestx}. Each image may have multiple labels.
\textbf{(4) Shenzhen dataset} is a small chest X-ray dataset with 662 frontal chest X-rays, of which 326 are normal cases and 336 are cases with manifestations of Tuberculosis \cite{jaeger2014two}.

\noindent\textbf{Training Details.}
We use Resnet18 as the backbone. The online projector and predictor follow the classical BYOL \cite{grill2020bootstrap}, and the embedding dimension is set to 256. On both ISIC 2019 and ChestX-ray datasets, we resize all the images to 140$\times$140 and then crop them to 128$\times$128. Data augmentation used in pre-training includes horizontal flipping, vertical flipping, rotation, color jitter, and cropping. For TracIn computation, we use one view with no augmentation and the other view with horizontal flipping and center cropping because this setting has the best empirical results in our experiments. We pre-train the model for 300 epochs using SGD optimizer with momentum 0.9 and weight decay $1\times e^{-5}$. The learning rate is set to 0.1 for the first 10 epochs and then decays following a concise learning rate schedule. The batch size is set to 256. The moving average decay of the momentum encoder is set to 0.99 at the beginning and then gradually updates to 1 following a concise schedule. 
All experiments are performed on one NVIDIA GeForce GTX 1080 GPU.

\noindent\textbf{Baselines.}
We compare the performance of our method with a random initialization approach without pre-training and the following SOTA baselines that involve pre-training. 
(1) BYOL \cite{grill2020bootstrap}: the vanilla BYOL with one positive pair from the same image. (2) FNC \cite{huynh2022boosting}: a false negative identification method designed to improve contrastive-based SSL framework. We adapt it to BYOL to select additional positives because false negatives are also equal to true positives for a particular anchor sample. (3) FT \cite{zhu2021improving}: a feature transformation method used in contrastive learning that creates harder positives and negatives to improve the learning ability. We apply it in BYOL to create harder virtual positives. (4) FS: using feature similarity from the current mini-batch to select the top-1 additional positive. (5) FS-pretrained: different from the FS that uses the current model to compute the feature similarity on the fly, we use a pre-trained model to test whether a well-trained encoder is more helpful in identifying the additional positives. (6) BYOL-Sup: the supervised BYOL in which we randomly select one additional positive from the mini-batch using the label information. This baseline is induced as the upper bound of our method because the additional positive is already correct.
We evaluate two variants of our method, BYOL-TracIn and BYOL-TracIn-pretrained. The former uses the current training model to compute the TracIn for each iteration while the latter uses a pre-trained model.
For a fair comparison, all methods use the same pre-training and finetuning setting unless otherwise specified. For FS-pretrained and BYOL-TracIn-pretrained, the pre-trained model uses the same setting as BYOL. 
Note that this pre-trained model is only used for positive selection and not involves in training.

\subsection{Semi-supervised Learning}

\begin{table}[t]
\centering
\caption{Comparison of all methods on ISIC 2019 and ChestX-ray datasets in the semi-supervised setting. We also report the fine-tuning results on 100$\%$ datasets. BYOL-Sup is the upper bound of our method. BMA represents the balanced multiclass accuracy.}
\resizebox{\columnwidth}{!}{
\begin{tabular}{l|ccc|ccc}
\toprule
& & ISIC 2019 & & & ChestX-ray &  \\
Method & 10$\%$ & 50$\%$ & 100$\%$ & 10$\%$ & 50$\%$ & 100$\%$ \\ 
& & BMA $\uparrow$ & & & AUC $\uparrow$ & \\
\hline 
Random & 0.327(.004) & 0.558(.005) & 0.650(.004) & 0.694(.005) & 0.736(.001) & 0.749(.001) \\
BYOL \cite{grill2020bootstrap} & 0.399(.001) & 0.580(.006) & 0.692(.005) & 0.699(.004) & 0.738(.003) & 0.750(.001) \\
FNC \cite{huynh2022boosting} & 0.401(.004) & 0.584(.004) & 0.694(.005) & 0.706(.001) & 0.739(.001) & 0.752(.002) \\
FT \cite{zhu2021improving} & 0.405(.005) & 0.588(.008) & 0.695(.005) & 0.708(.001) & 0.743(.001) & 0.751(.002) \\
FS & 0.403(.006) & 0.591(.003) & 0.694(.004) & 0.705(.003) & 0.738(.001) & 0.752(.002) \\
FS-pretrained & 0.406(.002) & 0.596(.004) & 0.697(.005) & 0.709(.001) & 0.744(.002) & 0.752(.002) \\ \hline
BYOL-TracIn & 0.403(.003) & 0.594(.004) & 0.694(.004) & 0.705(.001) & 0.742(.003) & 0.753(.002) \\
\makecell[l]{BYOL-TracIn\\-pretrained} & \textbf{0.408(.007)} & \textbf{0.602(.003)} & \textbf{0.700(.006)} & \textbf{0.712(.001)} & \textbf{0.746(.002)} & \textbf{0.754(.002)} \\ \hline
BYOL-Sup & 0.438(.006) & 0.608(.007) & 0.705(.005) & 0.714(.001) & 0.748(.001) & 0.756(.003) \\
\bottomrule
\end{tabular}
}
\label{table:semi_supervised}
\end{table}

In this section, we evaluate the performance of our method by finetuning with the pre-trained encoder on the same dataset as pre-training with limited annotations. We sample 10$\%$ or 50$\%$ of the labeled data from ISIC 2019 and ChestX-ray training sets and finetune the model for 100 epochs on the sampled datasets.
Data augmentation is the same as pre-training. Table \ref{table:semi_supervised} shows the comparisons of all methods. For ISIC 2019, we report the balanced multiclass accuracy (BMA, suggested by the ISIC challenge). For ChestX-ray, we report the average AUC across all diagnoses. We conduct each finetuning experiment 5 times with different random seeds and report the mean and std.

From Table \ref{table:semi_supervised}, we have the following observations: (1) Compared to Random, all the other methods have better accuracy, which means that pre-training can indeed help downstream tasks. (2) Compared to vanilla BYOL, other pre-training methods show performance improvement on both datasets. 
This shows that additional positives can increase feature diversity and benefit BYOL learning. (3) Our BYOL-TracIn-pretrained consistently outperforms all other unsupervised baselines. Although BYOL-TracIn can improve BYOL, it could be worse than other baselines like FT and FS-pretrained (e.g., 10$\%$ on ISIC 2019). This is because some additional positives identified by the on-the-fly model may be false positives, 
and attracting representations of such samples will degrade the learned features. 
However, with a pre-trained model in BYOL-TracIn-pretrained, the identification accuracy can be increased, leading to more true positives and better representations. (4) TracIn-pretrained performs better than FS-pretrained in all settings, and the improvement in BMA could be up to 0.006. This suggests that TracIn can be a more reliable metric for assessing the similarity between images when there is no human label information available. (5) Supervised BYOL can greatly increase the BYOL performance on both datasets. Yet our BYOL-TracIn-pretrained only has a marginal accuracy drop from supervised BYOL with a sufficient number of training samples (e.g., 100$\%$ on ISIC 2019).

\begin{figure}[t]
\begin{floatrow}
\ffigbox{%
\includegraphics[width=0.49\textwidth]{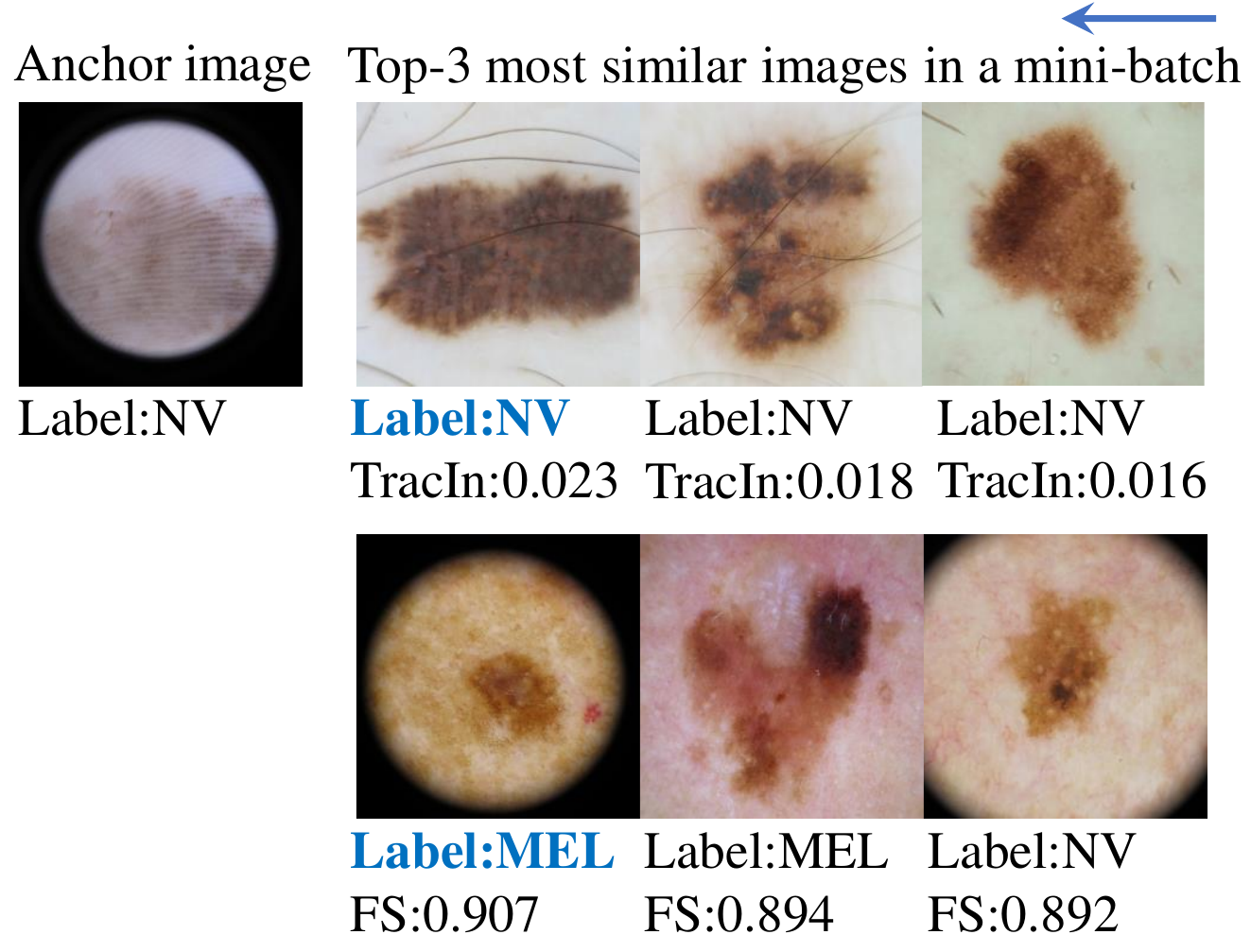}
  \vspace{-15pt}
}
{%
  \caption{Comparison of TracIn and Feature Similarity (FS) in selecting the additional positive during training on ISIC 2019.}%
  \label{figure:tracin_vs_fs}}
\capbtabbox{%
    \renewcommand{\arraystretch}{0.84}
    \begin{tabular}{l|c|c}
    \toprule
     Method &  ISIC 2016 & Shenzhen\\ 
      &  Precision $\uparrow$ & AUC $\uparrow$ \\ 
    \hline 
    Random & 0.400(.005) & 0.835(.010) \\
    BYOL \cite{grill2020bootstrap} & 0.541(.008) & 0.858(.003) \\
    FNC \cite{huynh2022boosting} & 0.542(.007) & 0.862(.006) \\
    FT \cite{zhu2021improving} & 0.559(.011) & 0.876(.005) \\
    FS & 0.551(.003) & 0.877(.004) \\
    FS-pretrained & 0.556(.004) & 0.877(.006) \\ \hline
    BYOL-TracIn & 0.555(.012) & 0.880(.007) \\
    \makecell[l]{BYOL-TracIn\\-pretrained} & \textbf{0.565(.010)} & \textbf{0.883(.001)} \\ \hline
    BYOL-Sup & 0.592(.008) & 0.893(.006) \\
    \bottomrule
    \end{tabular}
}{%
  \caption{Transfer learning comparison of the proposed method with the baselines on ISIC 2016 and Shenzhen datasets.}%
  \label{table:transfer_learning}
}
\end{floatrow}
\end{figure}

To further demonstrate the superiority of TracIn over Feature Similarity (FS) in selecting additional positive pairs for BYOL, we use an image from ISIC 2019 as an example and visualize the top-3 most similar images selected by both metrics using a BYOL pre-trained model in Fig. \ref{figure:tracin_vs_fs}. We can observe that TracIn accurately identifies the most similar images with the same label as the anchor image, whereas two of the images selected by FS have different labels. This discrepancy may be attributed to the fact that the FS of these two images is dominated by unrelated features (e.g., background tissue), which makes it unreliable. More visualization examples can be found in the supplementary.

\subsection{Transfer Learning}
To evaluate the transfer learning performance of the learned features, we use the encoder learned from the pre-training to initialize the model on the downstream datasets (ISIC 2019 transfers to ISIC 2016, and ChestX-ray transfers to Shenzhen). We finetune the model for 50 epochs and report the precision and AUC on ISIC 2016 and Shenzhen datasets, respectively. Table \ref{table:transfer_learning} shows the comparison results of all methods. We can see that BYOL-TracIn-pretrained always outperforms other unsupervised pre-training baselines, indicating that the additional positives can help BYOL learn better transferrable features.



\section{Conclusion}
In this paper, we propose a simple yet effective method, named BYOL-TracIn, to boost the representation learning performance of the vanilla BYOL framework. BYOL-TracIn can effectively identify additional positives from different samples in the mini-batch without using label information, thus introducing more variances to learned features. Experimental results on multiple public medical image datasets show that our method can significantly improve classification performance in both semi-supervised and transfer learning settings.


%
%
%
\bibliographystyle{splncs04}
\bibliography{reference.bib}
\end{document}


\title{Supplementary: Additional Positive Enables Better Representation Learning for Medical Images}

\author{}
\institute{}

\maketitle             

\begin{figure}[ht]
\centering
  \includegraphics[width=0.8\textwidth]{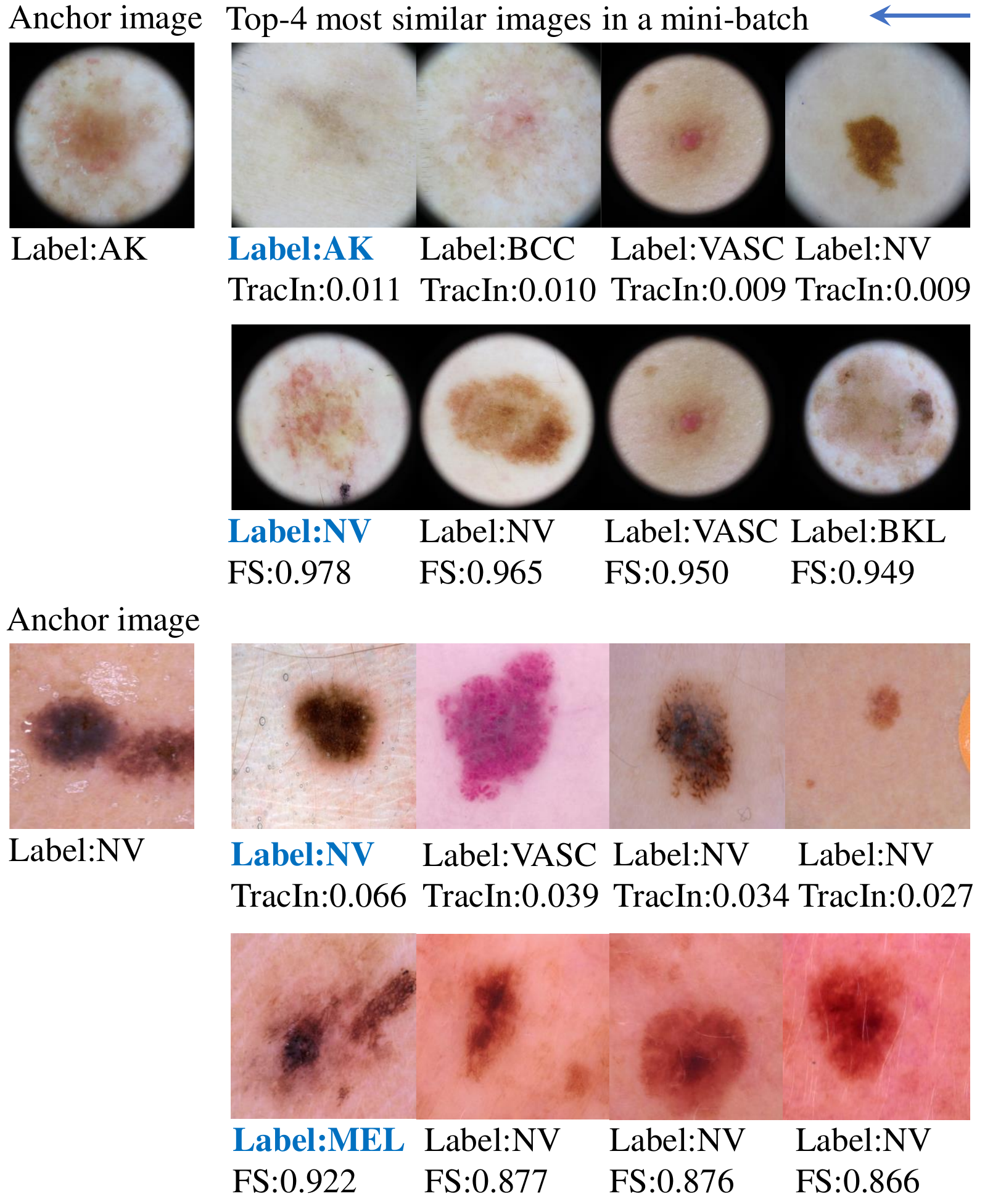}
\caption{Comparison of TracIn and Feature Similarity (FS) in selecting additional positive pairs. For each anchor sample from ISIC 2019, we visualize the top-4 most similar images selected by TracIn and FS. We can see that FS might select false positives as the additional samples whereas TracIn could identify the correct images with the same label as the anchor samples.}
\label{fig:framework_scl}
\end{figure}